\documentclass[runningheads]{llncs}
\usepackage{makeidx}
\usepackage{graphicx}
\usepackage{amsmath,amssymb} 
\usepackage{color,xcolor}
\usepackage{xspace}
\usepackage{subfigure}
\usepackage{epsfig}%
\usepackage{multirow} 
\usepackage{slashbox}
\usepackage{todonotes}
\usepackage{array}
\usepackage{color}
\usepackage{float}
\usepackage[normalem]{ulem}

\usepackage{caption}

\newcommand{\etal}{et al.\xspace} 

\begin{document}

    \pagestyle{headings}
    \mainmatter
    
    
    \title{End-to-End Learning of Video Super-Resolution with Motion Compensation}

	\titlerunning{End-to-End Learning of Video Super-Resolution with Motion Compensation}
	\authorrunning{Osama Makansi, Eddy Ilg and Thomas Brox}
	\author{Osama Makansi, Eddy Ilg, and Thomas Brox}
	\institute{Department of Computer Science, University of Freiburg}

	\maketitle

	\begin{abstract}
    Learning approaches have shown great success in the task of super-resolving an image given a low resolution input. Video super-resolution aims for exploiting additionally the information from multiple images. Typically, the images are related via optical flow and consecutive image warping. In this paper, we provide an end-to-end video super-resolution network that, in contrast to previous works, includes the estimation of optical flow in the overall network architecture. 
	We analyze the usage of optical flow for video super-resolution and find that common off-the-shelf image warping does not allow video super-resolution to benefit much from optical flow. 
	We rather propose an operation for motion compensation that performs warping from low to high resolution directly. We show that with this network configuration, video super-resolution can benefit from optical flow and we obtain state-of-the-art results on the popular test sets. 
	We also show that the processing of whole images rather than independent patches is responsible for a large increase in accuracy. 	    
	\end{abstract}

	\section{Introduction}
	\label{sec:introduction}
	
	The task of providing a good estimation of a high-resolution (HR) image from low-resolution (LR) input with minimum upsampling effects, such as ringing, noise, and blurring has been studied extensively~\cite{free02,chang04,yang08,yang10}. In recent years, deep learning approaches have led to a significant increase in performance on the task of image super-resolution~\cite{dong16,kim16,kim16_2,ledig16}.  
	Potentially, multiple frames of a video provide extra information that allows even higher quality up-sampling than just a single frame. 
	However, the task of simultaneously super-resolving multiple frames is inherently harder and thus has not been investigated as extensively. The key difficulty from a learning perspective is to relate the structures from multiple frames in order to assemble their information to a new image. 
	
	
    Kappeler~\etal~\cite{kapp16} were the first who proposed a convolutional network (CNN) for video super-resolution. They excluded the frame registration from the learning problem and rather applied motion compensation (warping) of the involved frames using precomputed optical flow. Thus, only a small part of the video super-resolution task was learned by the network, whereas large parts of the problem rely on classical techniques. 
    
    \pagebreak
    In this work, we provide for the first time an end-to-end network for video super-resolution that combines motion compensation and super-resolution into a single network with fast processing time. To this end, we make use of the FlowNet2-SD for optical flow estimation~\cite{ilg17}, integrate it into the approach by Kappeler~\etal~\cite{kapp16}, and train the joint network end-to-end. The integration requires changing the patch-based training~\cite{dong16,kapp16} to an image-based training and we show that this has a positive effect.   
    We analyze the resulting approach and the one from Kappeler~\etal ~\cite{kapp16} on single, multiple, and multiple motion-compensated frames in order to quantify the effect of using multiple frames and the effect of motion estimation. The evaluation reveals that with the original approach from Kappeler~\etal both effects are surprisingly small. Contrary, when switching to image-based trainng we see an improvement 
    if using motion compensated frames and we obtain the best results with the \mbox{FlowNet2-SD} motion compensation.
        
    The approach of Kappeler~\etal~\cite{kapp16} follows the common practice of first upsampling and then warping images. Both operations involve an interpolation by which high-frequency image information is lost. To avoid this effect, we then implement a motion compensation operation to directly perform upsampling and warping in a single step. We compare to the closely related work of Tao et al.~\cite{tao17} and also perform experiments with their network architecture. Finally, we show that with this configuration, CNNs for video super-resolution clearly benefit from optical flow. We obtain state-of-the-art results.

    \section{Related work}      
    
    \subsection{Image super-resolution}
    The pioneering work in super-resolving a LR image dates back to Freeman~\etal~\cite{free02}, who used a database of LR/HR patch examples and nearest neighbor search to perform restoration of a HR image. 
    Chang~\etal~\cite{chang04} replaced the nearest neighbor search by a manifold embedding, while Yang~\etal built upon sparse coding~\cite{yang08,yang10}.
    Dong~\etal~\cite{dong16} proposed a convolutional neural network (SRCNN) for image super-resolution. They introduced an architecture consisting of the three steps patch encoding, non-linear mapping, and reconstruction, and showed that CNNs outperform previous methods.
    In Dong~\etal~\cite{dong16_2}, the three-layer network was replaced by a convolutional encoder-decoder network with improved speed and accuracy. Shi~\etal~\cite{shi16} showed that performance can be increased by computing features in the lower resolution space. Recent work has extended SRCNN to deeper~\cite{kim16} and recursive~\cite{kim16_2} architectures. Ledig~\etal~\cite{ledig16} employed generative adversarial networks.
    
    \subsection{Video super-resolution}

    Performing super-resolution from multiple frames is a much harder task due to the additional alignment problem. Many approaches impose restrictions, such as the presence of HR keyframes~\cite{song11} or affine motion~\cite{baba11}. Only few general approaches exist. Liu~\etal~\cite{liu14} provided the most extensive approach by using a Bayesian framework to estimate motion, camera blur kernel, noise level, and HR frames jointly. Ma~\etal~\cite{ma15} extended this work to incorporate motion blur. 
    Takeda~\etal~\cite{takeda09} followed an alternative approach by considering the video as a 3D spatio-temporal volume and by applying multidimensional kernel regression. 
    
    A first learning approach to the problem was presented by Cheng~\etal~\cite{hui12}, who used block matching to find corresponding patches and applied a multi-layer perceptron to map the LR spatio-temporal patch volumes to HR pixels. Kappeler~\etal~\cite{kapp16} proposed a basic CNN approach for video-super-resolution by extending SRCNN to multiple frames. Given the LR input frames and optical flow (obtained with the method from~\cite{drulea11}), they bicubically upsample and warp distant time frames to the current one and then apply a slightly modified SRCNN architecture (called VSR) on this stack. The motion estimation and motion compensation are provided externally and are not part of the training procedure.
    
    Caballero~\etal~\cite{jose16} proposed a spatio-temporal network with 3D convolutions and slow fusion to perform video super-resolution. They employ a multi-scale spatial transformer module for motion compensation, which they train jointly with the 3D network. Very recently, Tao~\etal~\cite{tao17} used the same motion compensation transformer module. Instead of a 3D network, they proposed a recurrent network with an LSTM unit to process multiple frames. Their work introduces an operation they call SubPixel Motion Compensation (SPMC), which performs forward warping and upsampling jointly. This is strongly related to the operation we propose here, though we use backward warping combined with a confidence instead of forward warping. Moreover, we use a simple feed-forward network instead of a recurrent network with an LSTM unit, which is advantageous for training. 
    
    \subsection{Motion estimation} 
    
    Motion estimation is a longstanding research topic in computer vision, and a survey is given in~\cite{sun10}. In this work, we aim to perform video super-resolution with a CNN-only approach. The pioneering FlowNet of Dosovitskiy~\etal~\cite{doso15} showed that motion estimation can be learned end-to-end with a CNN. Later works~\cite{ranjan16,ilg17} elaborated on this concept and provided multiscale and multistep approaches. The FlowNet2 by Ilg~\etal~\cite{ilg17} yields state-of-the-art accuracy but is orders of magnitudes faster than traditional methods. We use this network as a building block for end-to-end training of a video super-resolution network. 

    \section{Video super-resolution with patch-based training} 
    
\begin{figure}[H]
        \subfigure[\label{fig:vsr_architecture}Architecture as proposed by Kappeler~\etal \cite{kapp16}]{\includegraphics[width=\linewidth]{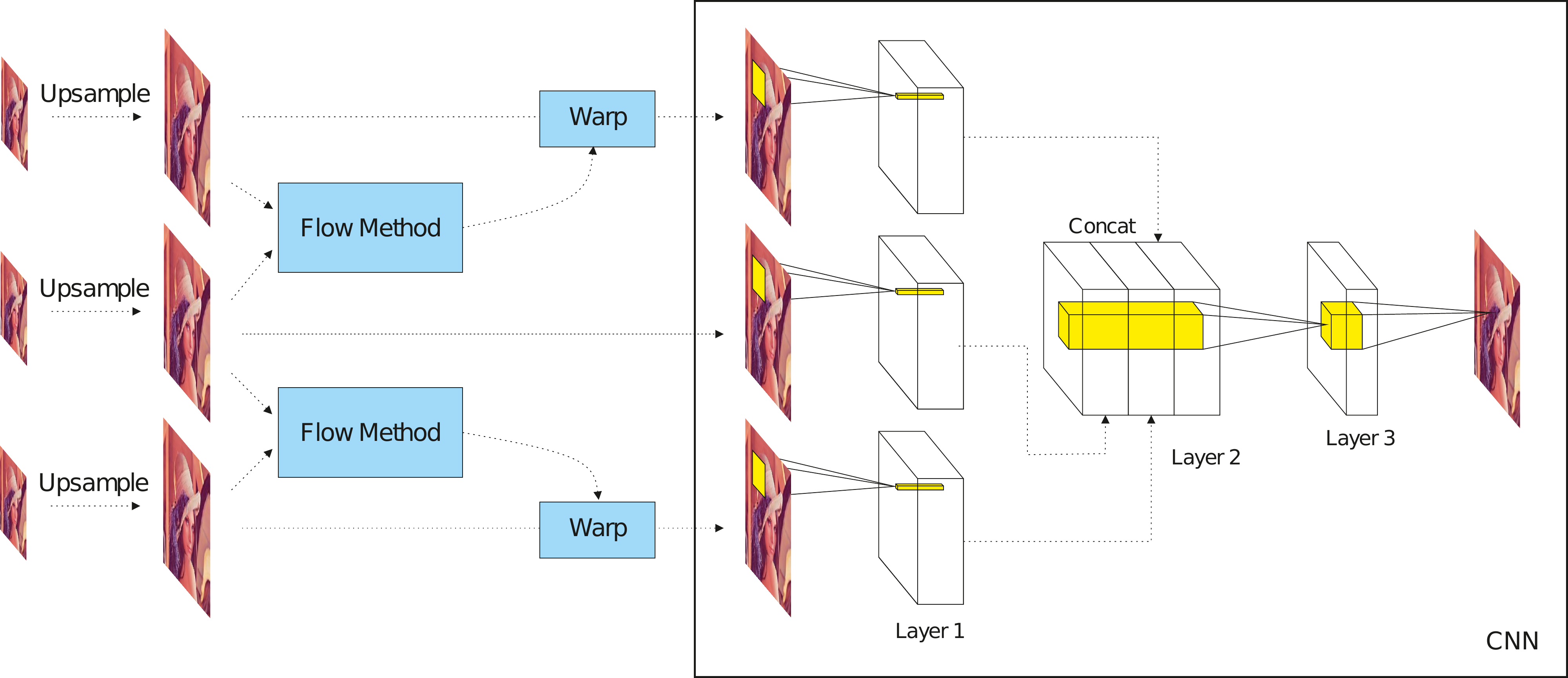}}
        \subfigure[\label{fig:vsr_joint_architecture}Architecture with integrated FlowNet2-SD from~\cite{ilg17}]{\includegraphics[width=\linewidth]{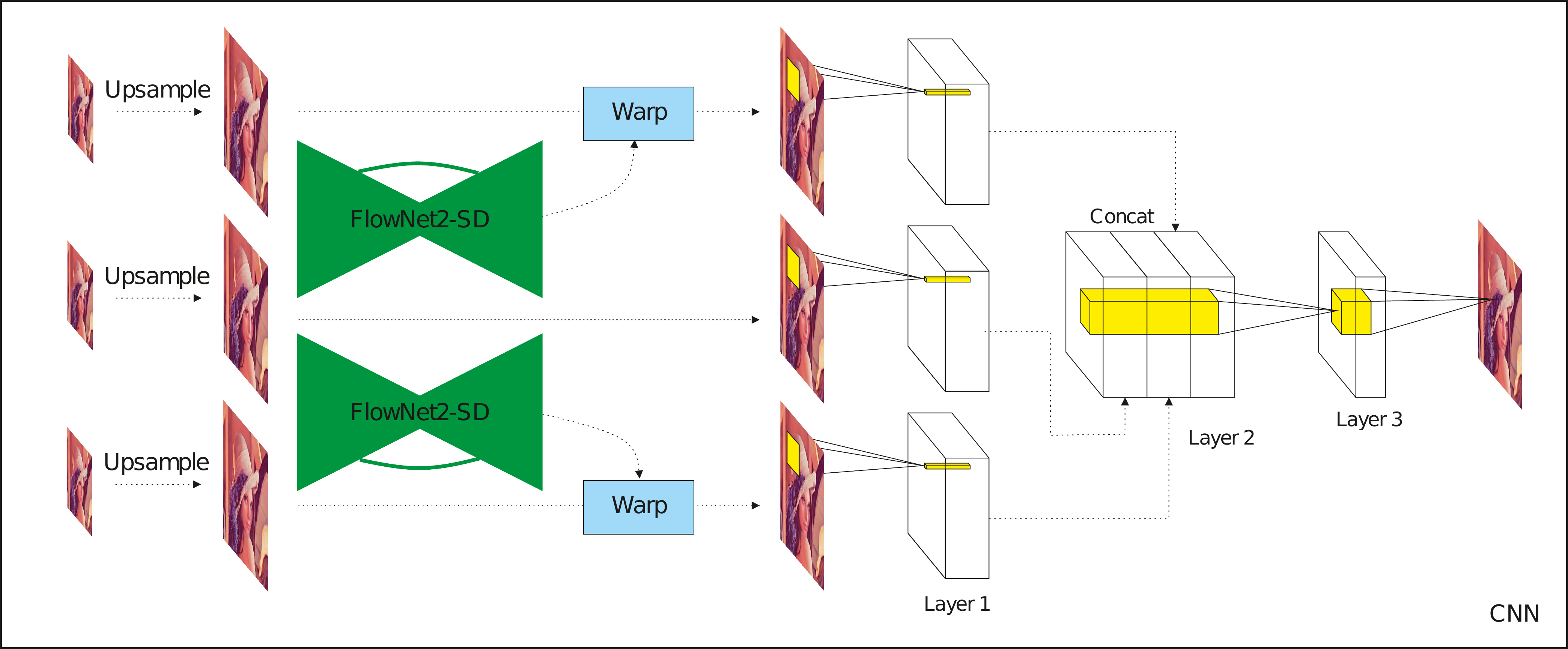}}
        \caption{Video super-resolution architectures used by the basic models tested in this paper. Optical flow is estimated from the center to the outer frames using either an external method or a CNN. The flow is used to warp all frames to the center frame. The frames are then input into to the VSR network~\cite{kapp16}. The complete network in (b) can be trained end-to-end including the motion estimation.}
        \vspace*{-2mm}
    \end{figure}

    In this section we revisit the work from Kappeler~\etal~\cite{kapp16}, which applies network-external motion compensation and then extends the single-image SRCNN~\cite{dong16} to operate on multiple frames. This approach is shown in Figure~\ref{fig:vsr_architecture}.
    \noindent
    Kappeler~\etal~\cite{kapp16} compare different numbers of input frames and investigate early and late fusion by
    performing the concatenation of features from the different frames after different layers. They conclude that fusion after the first convolution works best. 
    Here, we use this version and furthermore stick to three input frames and an upsampling factor of four throughout the whole paper.     
        
    
    We performed an analysis of their code and model. The results are given in the first row of Table~\ref{tab:orig_code_results}. 
    Using their original code, we conducted an experiment, where we replaced the three frames from the image
    sequence by three times the same center frame (column 4 of Table~\ref{tab:orig_code_results}), which corresponds     to the information only from single-image super-resolution. We find that on the Myanmar validation set the 
result is still much better than \mbox{SRCNN}~\cite{dong16} but only marginally worse than VSR~\cite{kapp16} on real video information. Since except for a concatenation there is no difference between the VSR~\cite{kapp16} and SRCNN~\cite{dong16} architectures, this shows that surprisingly the improvement is mainly due to training settings of VSR~\cite{kapp16} rather than the usage of multiple frames. 
    
    For training and evaluation, Kappeler~\etal~\cite{kapp16} used the publicly available Myanmar video~\cite{myanmardata}. We used the same training/validation split into 53 and 6 scenes and followed the patch sampling from~\cite{kapp16}. However, the publicly available data has changed by that the overlaid logo at the bottom right corner from the producing company is now bigger than before.
    Evaluating on the data with the different logo gives much worse results (row 2 of Table~\ref{tab:orig_code_results}), while when the logo is cropped off (column 3 of Table~\ref{tab:orig_code_results}), results are comparable. The remaining difference stems from a different implementation of the warping operation\footnote{We use the implementation from~\cite{ilg17}; it differs from~\cite{kapp16} in that it performs bilinear interpolation instead of bicubic.}. However, when we retrained the approach with our implementation and training data (row 3 of Table~\ref{tab:orig_code_results}), we achieved results very close to Kappler et al.~\cite{kapp16}. 

    To further investigate the effects of motion compensation, we retrained the approach using only the center frame, the original frames, and frames motion compensated using FlowNet2~\cite{ilg17} and FlowNet2-SD~\cite{ilg17} in addition to the method from Drulea~\cite{drulea11}. For details we refer to the supplemental material. 
    Again we observed that including or excluding motion compensation with different optical flow methods has no effect on the Myanmar validation set. We additionally evaluated on the commonly used Videoset4 dataset \cite{liu14,kapp16}. In this case we do see a PSNR increment of $0.1$ with Drulea~\cite{drulea11} and higher increment of $0.18$ with FlowNet2~\cite{ilg17} when using motion compensation. The Videoset4 dataset includes larger motion and it seems that there is some small improvement when larger motion is involved. However, the effect of motion compensation is still very small when compared to the effect of changing other training settings.
    
    \section{Video super-resolution with image-based training\label{sec:image-based}}
    
    In contrast to Kappeler~\etal, we combine motion compensation and super-resolution in one network. For motion estimation, we used the FlowNet2-SD variant from \cite{ilg17}. We chose this network, because FlowNet2 itself is too large to fit into GPU memory besides the super-resolution network and FlowNet2-SD yields smooth flow predictions and accurate performance for small displacements. Figure~\ref{fig:vsr_joint_architecture} shows the integrated network. For the warping operation, we use the implementation from~\cite{ilg17}, which also allows a backward pass while training. 
    The combined network is trained on complete images instead of patches. Thus, we repeated our experiments from the previous section for the case of image-based training. The results are given in  Table~\ref{tab:image_and_joint}. In general, we find that image-based processing yields much higher PSNRs than patch-based processing. Detailed comparison of the network and training settings for both variants can be found in the supplemental material. 
    
        \begin{table}[H]
        \begin{center}
            \resizebox{\linewidth}{!}{
            \begin{tabular}{|l|@{}p{2pt}@{}|c|@{}p{2pt}@{}|c|c|c|c|} 
             \cline{1-1}\cline{3-3}\cline{5-8}       
             \multirow{2}{*}{Dataset/Model}
              && \multirow{2}{*}{SRCNN \cite{dong16}}  && \multirow{2}{*}{VSR \cite{kapp16}} & VSR \cite{kapp16} & VSR \cite{kapp16}          & VSR \cite{kapp16}\\
             
              &&       &&             & (cropped) & (only center) & (no warp.)\\
             \cline{1-1}\cline{3-3}\cline{5-8}
             \multicolumn{8}{c}{} \\[-0.8\normalbaselineskip]
             \cline{1-1}\cline{3-3}\cline{5-8}
             Myanmar validation from \cite{kapp16} && $31.26$ && $\textbf{31.81}$ & $32.95$ & $\textbf{31.71}$ & -\\
             \cline{1-1}\cline{3-3}\cline{5-8}
             \multicolumn{8}{c}{} \\[-0.8\normalbaselineskip]
             \cline{1-1}\cline{3-3}\cline{5-8}
             Myanmar validation (ours)             && \multirow{2}{*}{$31.30$} && $\textbf{31.30}$ & $32.88$ & $31.23$ & $31.19$\\
             Myanmar validation (ours), retrained  &&       && $\textbf{31.81}$ & $32.76$ &     $31.74$ & $31.77$ \\
             \cline{1-1}\cline{3-3}\cline{5-8}
            \end{tabular}
            }
            \vspace*{1mm}
            \caption{
                Analysis of Kappeler~\etal~\cite{kapp16} on the different versions of the Myanmar dataset. Numbers show the PSNR in dB.
                The first row is with the original code and test data from \cite{kapp16}, while the second and third row are with our re-implementation and the new test data that was recently downloaded. The third column shows results when the logo area is cropped off. Fourth and fifth columns show the PSNR when motion compensation is disabled during testing, by using only the center frame or the original frames without warping. There is no significant improvement by neither the use of multiple frames nor by the use of optical flow.   
            }
            \label{tab:orig_code_results}
        \end{center}
        \vspace*{-3mm}
    \end{table} 
    \begin{table}[H]
        \begin{center}
            \resizebox{\linewidth}{!}{
            \begin{tabular}{|l|@{}p{2pt}@{}|c|@{}p{2pt}@{}|c|c|c|c|c|}
                \cline{1-1}\cline{3-3}\cline{5-9}       
                Network && SRCNN \cite{dong16} && VSR \cite{kapp16} & VSR \cite{kapp16} & VSR \cite{kapp16} & VSR \cite{kapp16} & VSR \cite{kapp16} joint \\             
                \cline{1-1}\cline{3-3}\cline{5-9}       
                Motion Compensation && - && only center & no warp. & Drulea \cite{drulea11} & FN2-SD \cite{ilg17} & FN2-SD \cite{ilg17} \\                             
                \cline{1-1}\cline{3-3}\cline{5-9}       
                \multicolumn{9}{c}{} \\[-0.8\normalbaselineskip]
                \cline{1-1}\cline{3-3}\cline{5-9}       
                Myanmar validation (ours) && $32.42$ && $32.41$ & $32.55$ & $32.60$ & $32.62$ & $32.63$ \\
                Videoset4 && $24.63$ && $24.66$ & $24.79$ & $24.91$ & $25.12$ & $25.21$ \\
                \cline{1-1}\cline{3-3}\cline{5-9}       
            \end{tabular}
            }
            \vspace*{1mm}
            \caption{
            PSNR scores from  Myanmar validation (ours) and Videoset4 for image-based training. For each column of the table we trained the architecture of \cite{dong16} and \cite{kapp16} by applying convolutions over the complete images. We used different types of motion compensation for training and testing (FN2-SD denotes FlowNet2-SD). For Myanmar, motion compensation still has no significant effect. However, on Videoset4 an effect for motion compensation using Drulea's method~\cite{drulea11} is noticeable and is even stronger for FlowNet2-SD\cite{ilg17}.
            }
            \label{tab:image_and_joint}
        \end{center}
        \vspace*{-3mm}
    \end{table} 
    
    Table~\ref{tab:image_and_joint} shows that motion compensation has no effect on the Myanmar validation set. For Videoset4 there is an increase of $0.12$ with motion compensation using Drulea's method~\cite{drulea11}. For FlowNet2 the increase of $0.42$ is even bigger.
    Since FlowNet2-SD is completely trainable, it is also possible to refine the optical flow for the task of video super-resolution by training the whole network end-to-end with the super-resolution loss. We do so by using a resolution of $256\times256$ to enable a batch size of $8$ and train for $100$k more iterations. The results from Table~\ref{tab:image_and_joint} again show that for Myanmar there is no significant change. However, for Videoset4 the joint training further improves the result by $0.1$ leading to a total PSNR increase of $0.52$. 
 
\begin{figure} 
        \resizebox{\linewidth}{!}{
            \subfigure[ground truth]{\includegraphics[width=0.33\linewidth]{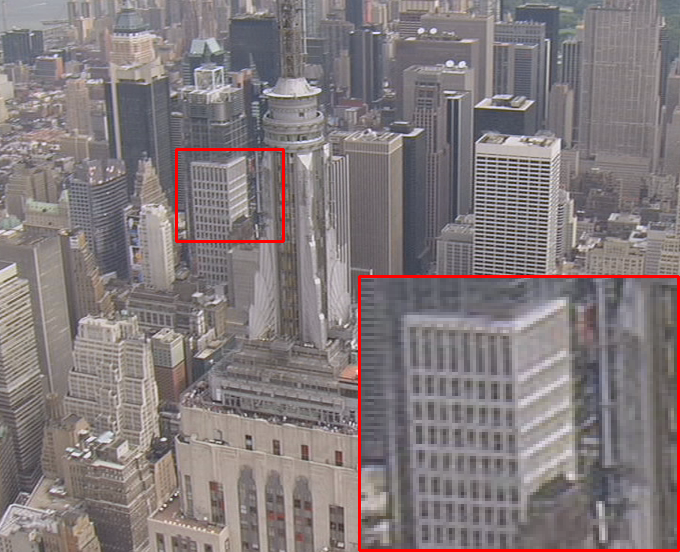}}
            \subfigure[SRCNN \cite{dong16}]{\includegraphics[width=0.33\linewidth]{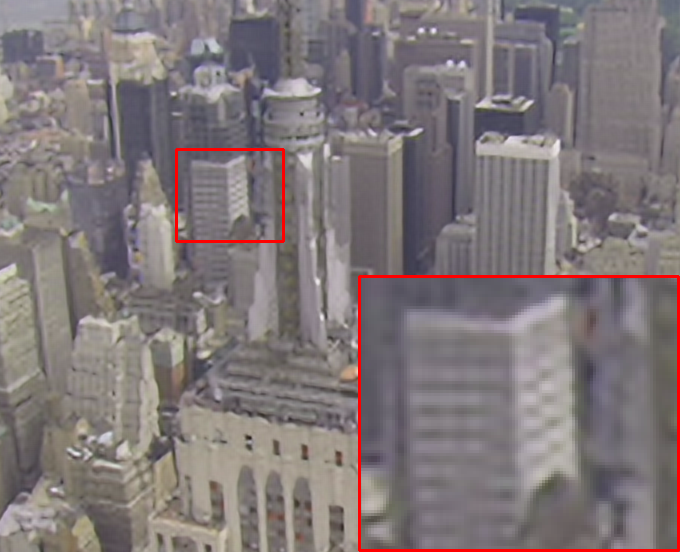}}
            \subfigure[VSR$^\dagger$ (only center)]{\includegraphics[width=0.33\linewidth]{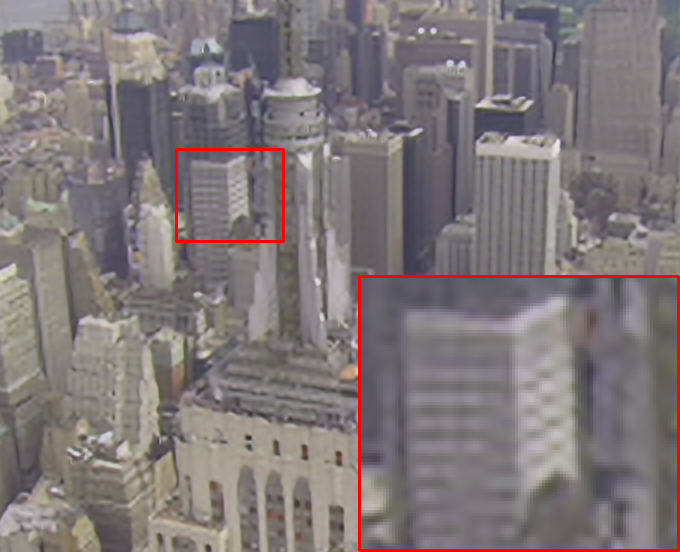}}
            \subfigure[VSR$^\dagger$ (Drulea \cite{drulea11})]{\includegraphics[width=0.33\linewidth]{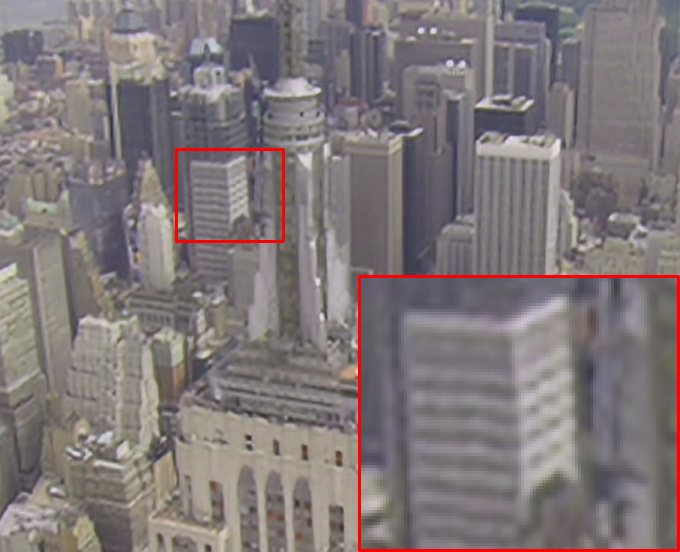}}
            \subfigure[Baysian \cite{liu14}]{\includegraphics[width=0.33\linewidth]{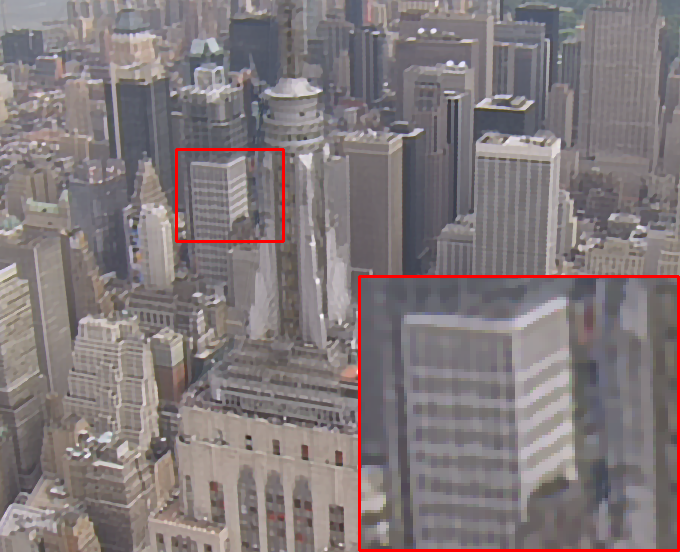}}
        }
        \resizebox{\linewidth}{!}{
            \subfigure[bicubic]{\includegraphics[width=0.33\linewidth]{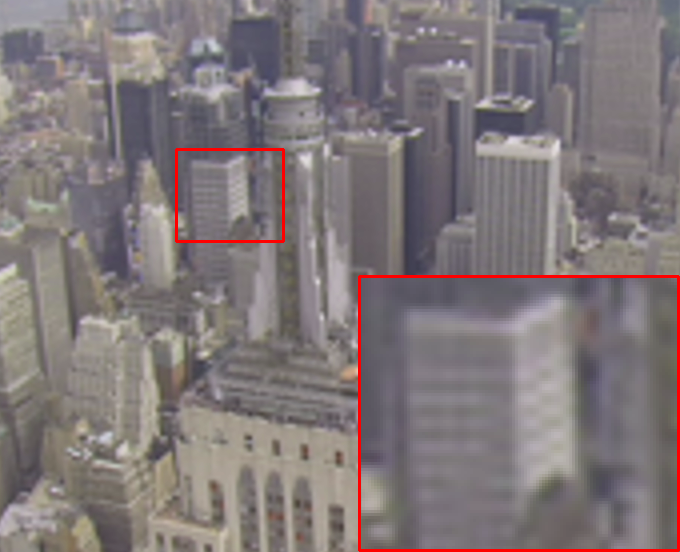}}
            \subfigure[VSR \cite{kapp16}]{\includegraphics[width=0.33\linewidth]{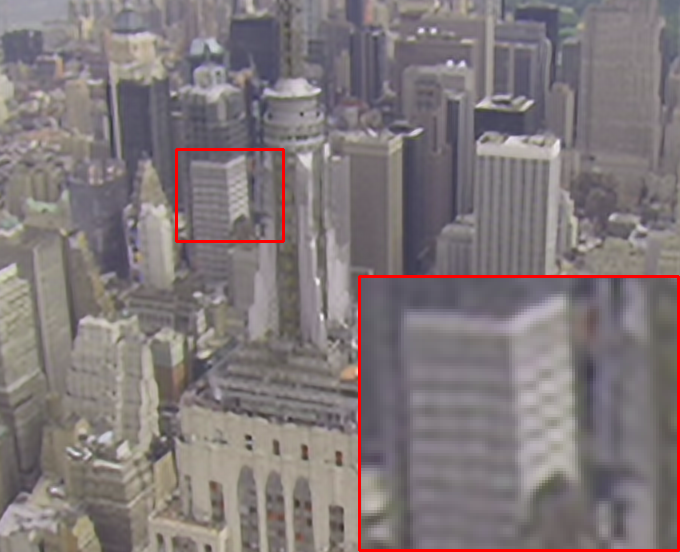}}
            \subfigure[VSR$^\dagger$ (no warp)]{\includegraphics[width=0.33\linewidth]{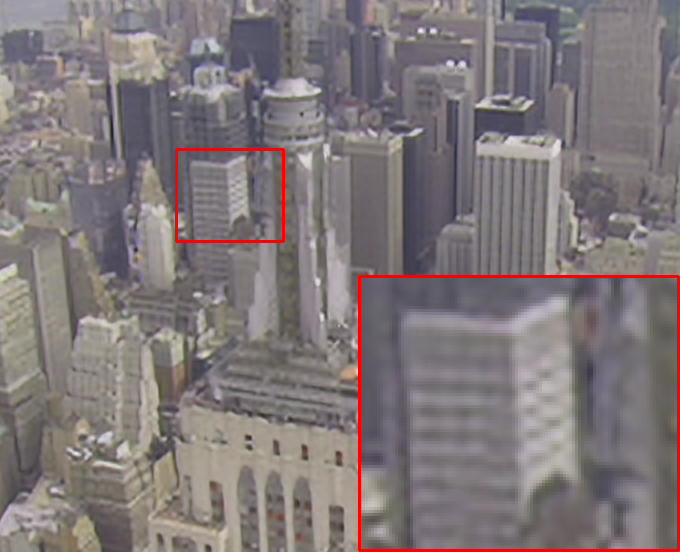}}
            \subfigure[VSR$^\dagger$ (FlowNet2-SD)]{\includegraphics[width=0.33\linewidth]{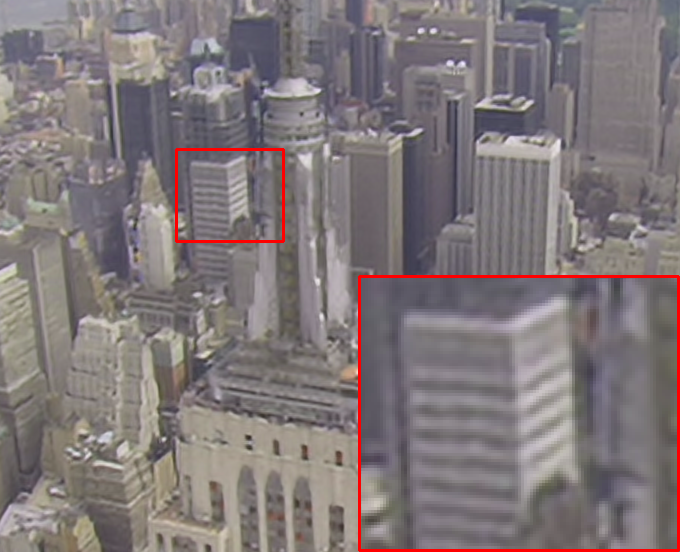}}
            \subfigure[VSR$^\dagger$ (FlowNet2-SD-joint)]{\includegraphics[width=0.33\linewidth]{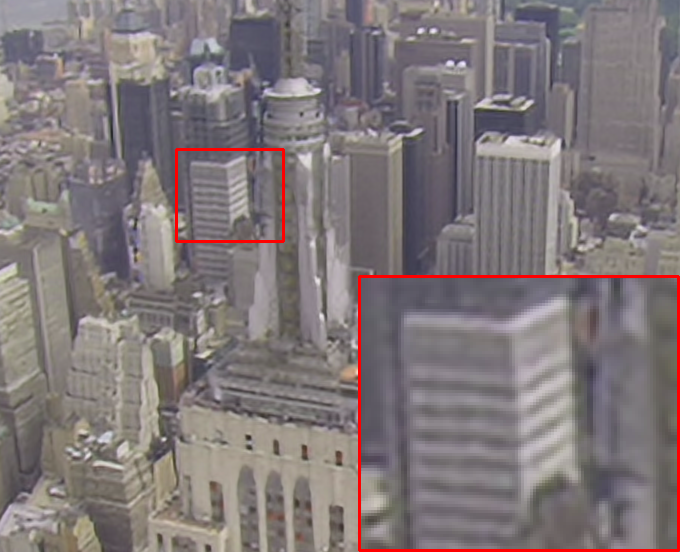}}
        }
        \caption{        
        Comparison of existing super-resolution methods to our trained models. 
        $^\dagger$ indicates models retrained by us using image-based training. Note that b) and g) are patch-based, while c), d), e), h), i) and j) are image-based.
        } 
        \label{fig:qualitative}
        \vspace*{-2mm}
    \end{figure}

    We show a qualitative evaluation in Figure~\ref{fig:qualitative}. On the enlarged building, one can see that bicubic upsampling introduces some smearing across the windows. This effect is also present in the methods without motion compensation and in the original VSR~\cite{kapp16} with motion compensation. When using image-based trained models, the effect is successfully removed. Motion compensation with FlowNet2~\cite{ilg17} seems to be marginally sharper than motion compensation with Drulea~\cite{drulea11}. We find that the joint training reduces ringing artifacts; an example is given in the supplemental material.

\section{Combined Warping and Upsampling Operation}

The approach of Kappeler~\etal~\cite{kapp16} and the VSR architecture discussed so far follow the common practice of first upsampling and then warping the images. Both operations involve an interpolation during which image information is lost.
Therefore, we propose a joint operation that performs upsampling and backward warping in a single step, which we name Joint Upsampling and Backward Warping (\emph{JUBW}). This operation does not perform any interpolation at all, but additionally outputs sub-pixel distances and leaves finding a meaningful interpolation to the network itself. 
Let us consider a pixel $p$ and let $x_p$ and $y_p$ denote the coordinates in high resolution space, while $x^{s}_p$ and $y^{s}_p$ denote the source coordinates in low resolution space. First, the mapping from low to high resolution space using high resolution flow estimations $(u_p, v_p)$ is computed according to the following equation:
\begin{equation}
\left( \begin{array}{c} x^{s}_p \\ y^{s}_p \end{array} \right) = \frac{1}{\alpha} \left( \begin{array}{c} x_p + u_p + 0.5 \\ y_p + v_p + 0.5 \end{array} \right) - \left( \begin{array}{c} 0.5 \\ 0.5 \end{array} \right) \mathrm{,}
\end{equation}

\begin{figure}
\begin{center}
\includegraphics[width=0.5\linewidth]{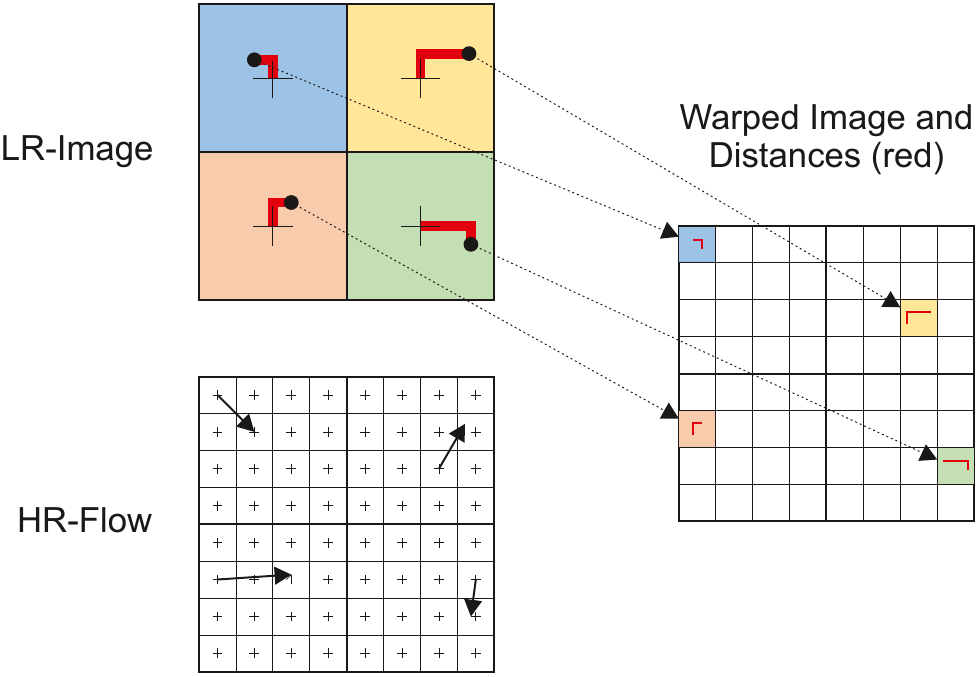}
\caption{
Illustration of the Joint Upsampling and Backward Warping operation (JUBW). The output is a dense image (left sparse here for illustration purposes) and includes $x$/$y$ distances of the source locations to the source pixel centers. 
\label{fig:bspmc_layer}
}
\end{center}
\end{figure}

\pagebreak 
\noindent
where $\alpha = 4$ denotes the scaling factor and subtraction/addition of $0.5$ places the origin at the top left corner of the first pixel. Then the warped image is computed as:
\begin{equation}
I_w(p)=
\begin{cases}
I(\left\lfloor x^{s}_p \right\rceil,\left\lfloor y^{s}_p \right\rceil) & \text{if } \left\lceil x^{s}_p \right\rfloor,\left\lceil y^{s}_p \right\rfloor \text{is inside $I$,} \\
0  & \text{otherwise} \mathrm{,}
\end{cases}
\end{equation}
where $\lfloor \cdot \rceil$ denotes the round to nearest operation. 
Note, that no interpolation between pixels is performed. The operation then additionally outputs the following distances per pixel (see Figure~\ref{fig:bspmc_layer} for illustration): 
\begin{equation}
\left( \begin{array}{c} d^{x}_p \\ d^{y}_p \end{array} \right) = 
\left( \begin{array}{c} \left\lfloor x^{s}_p \right\rceil - x^{s}_p \\  \left\lceil y^{s}_p \right\rceil - y^{s}_p \end{array} \right)
\text{if }  \left\lceil x^{s}_p \right\rfloor,\left\lceil y^{s}_p \right\rfloor \text{is inside $I$ and }
\left(\begin{array}{c} 0 \\ 0 \end{array}\right)
\text{otherwise.}
\end{equation}

\begin{figure}
\begin{center}
\includegraphics[width=\linewidth]{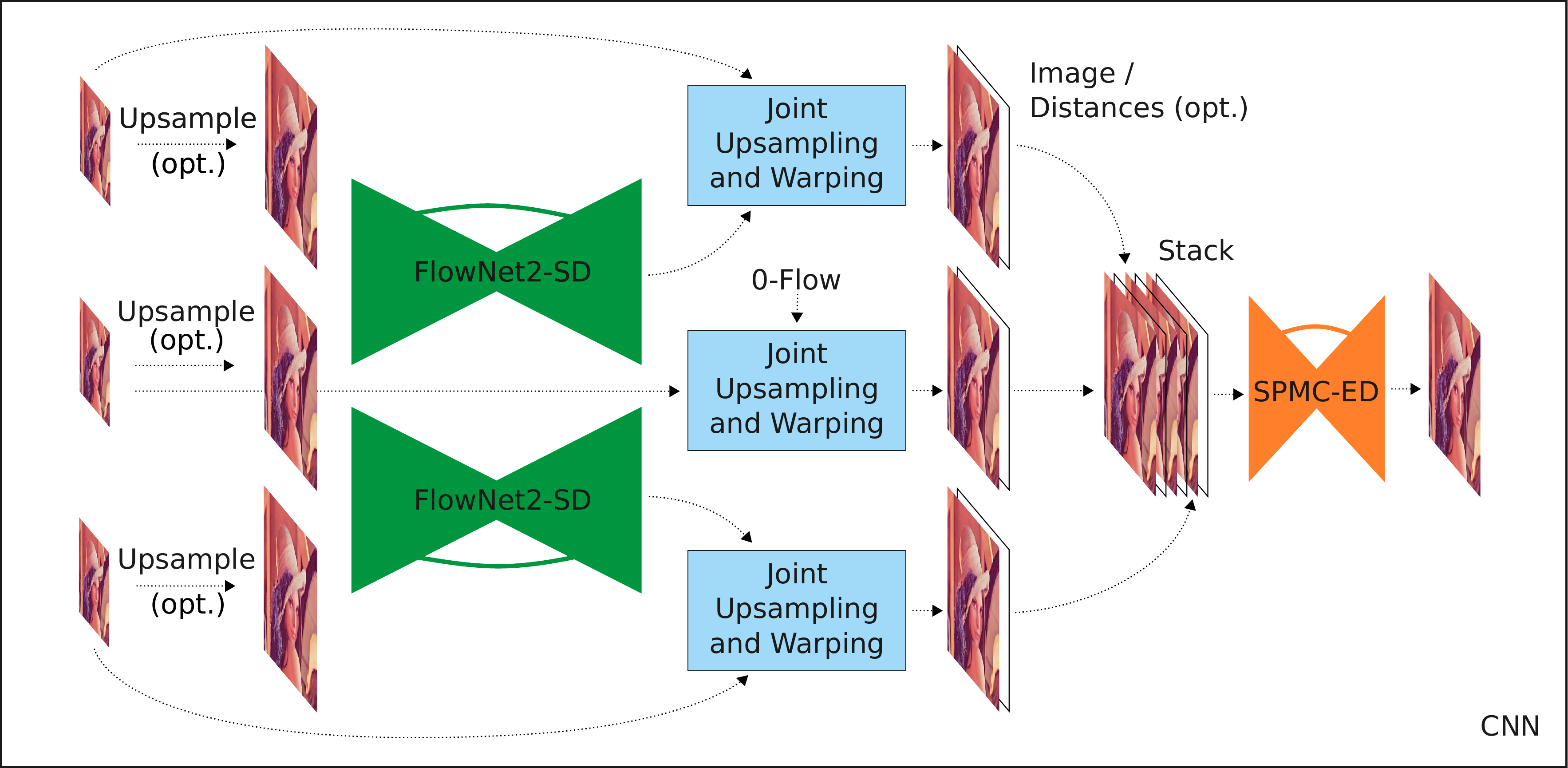}\vspace*{-2mm}
\caption{
Network setup with FlowNet2-SD and joint upsampling and warping operation (JUBW or SPMC-FW). Upsampling before feeding into FlowNet2-SD happens only for JUBW. The output of the upsampling and warping operation is stacked and then fed into the SPMC-ED network.  
\label{fig:jubw_setup}
}
\end{center}
\vspace*{-2mm}
\end{figure}

We also implemented the joint upsampling and forward warping operation from Tao~\etal~\cite{tao17} for comparison and denote it as SPMC-FW. 
Contrary to our operation, SMPC-FW still involves two types of interpolation: 1.) subpixel-interpolation for the target position in the high resolution grid and 2.) interpolation between values if multiple flow vectors point to the same target location.
For comparison, we replaced the architecture from the previous section by the \mbox{encoder-/decoder} part from Tao~\etal~\cite{tao17} (which we denote here as SPMC-ED). We also find that this architecture itself performs better than \mbox{SRCNN~\cite{dong16}/VSR~\cite{kapp16}}  on the super-resolution only task (see supplementary material for details). The resulting configuration is shown in Figure~\ref{fig:jubw_setup}. Furthermore, we also extended the training set by downloading Youtube videos and downsampling them to create additional training data. The larger dataset comprises 162k images and we call it MYT.

\begin{table}
    \centering
    \resizebox{\linewidth}{!}{
    
    \begin{tabular}{|c|@{}p{2pt}@{}|c|@{}p{2pt}@{}|c|c|c|@{}p{2pt}@{}|c|c|c|c|}
        \cline{3-3}\cline{5-7}\cline{9-12}
        \multicolumn{1}{c}{\multirow{2}{*}}&&
        \multicolumn{1}{c|}{SPMC}&&
        \multicolumn{3}{c|}{SPMC-FW}&&
        \multicolumn{4}{c|}{JUBW}\\ 
        \cline{5-7}\cline{9-12}
        \multicolumn{1}{c}{} && 
        original~\cite{tao17} && 
        ours & 
        only center & 
        joint && 
        ours & 
        no dist. & 
        only center & 
        joint\\ 
        \cline{3-3}\cline{5-7}\cline{9-12}
        \multicolumn{9}{c}{} \\[-0.8\normalbaselineskip]
        \cline{1-1}\cline{3-3}\cline{5-7}\cline{9-12}
        Myanmar (ours) && - && $32.90$ & $32.45$ & $33.05$ && $\textbf{33.13}$ & $33.02$ & $32.55$ & $32.69$ \\
        \cline{1-1}\cline{3-3}\cline{5-7}\cline{9-12}
        Videoset4 && $25.52$ && $25.68$ & $24.94$ & $25.62$ && $\textbf{25.85}$ & $25.74$ & $24.96$ & $25.09$ \\
        \cline{1-1}\cline{3-3}\cline{5-7}\cline{9-12}
    \end{tabular}
    }
    \vspace*{1mm}
    \caption{
    PSNR values for different joint upsampling and warping approaches. The first column shows the original results from Tao~\etal~\cite{tao17} using the SPMC upsampling, forward warping, and the SPMC-ED architecture with an LSTM unit. 
    Columns two to four show our reimplementation of the SPMC-FW operation~\cite{tao17} without an LSTM unit. Columns five to eight show our joint upsampling and backward warping operation with the same encoder-decoder network on top. With \emph{ours} we denote our implementation according to Figure~\ref{fig:jubw_setup}. In \emph{only center} we input zero-flows and the duplicated center image three times (no temporal information). The entry \emph{joint} includes joint training of FlowNet2-SD and the super-resolution network. For columns two to eight, the networks are retrained on MYT and tested for each setting respectively.          
    }

    \label{tab:warping_results} 
    \vspace*{-2mm}
\end{table}

\begin{figure}
    \resizebox{\linewidth}{!}{
        \subfigure[ground truth]{\includegraphics[width=0.33\linewidth]{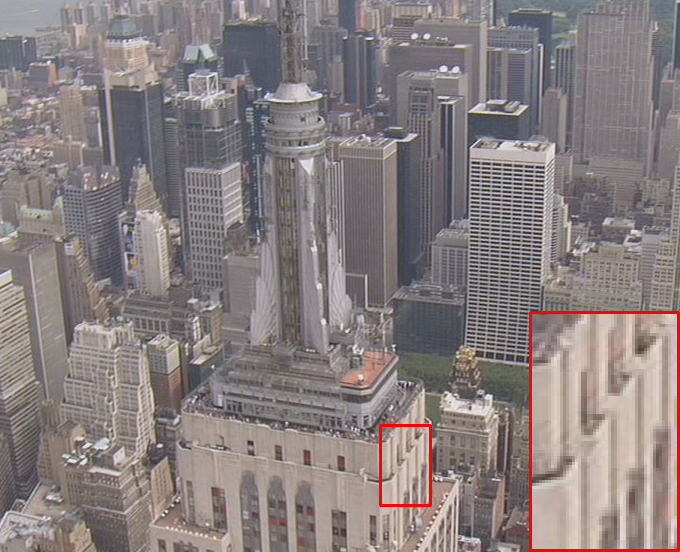}}
        \subfigure[FN2-SD+VSR joint]{\includegraphics[width=0.33\linewidth]{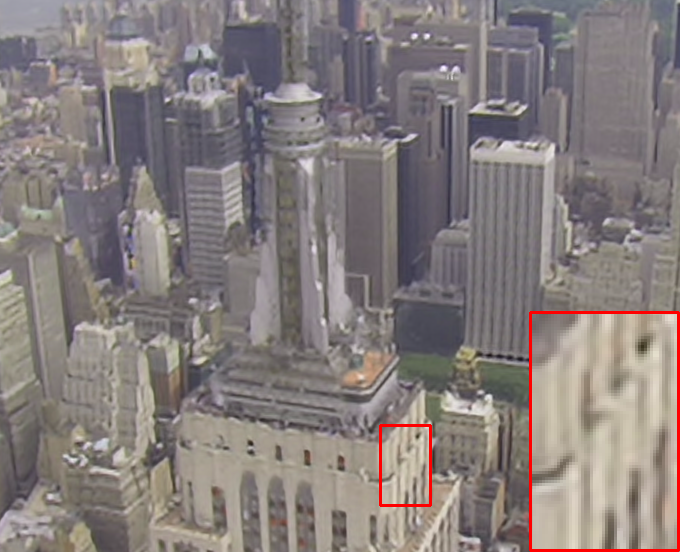}}
        \subfigure[FN2-SD+SPMC-FW]{\includegraphics[width=0.33\linewidth]{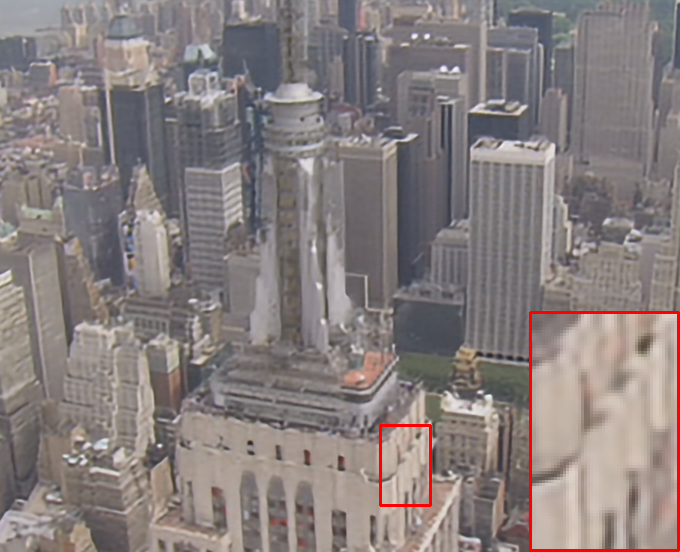}}
        \subfigure[FN2-SD+JUBW]{\includegraphics[width=0.33\linewidth]{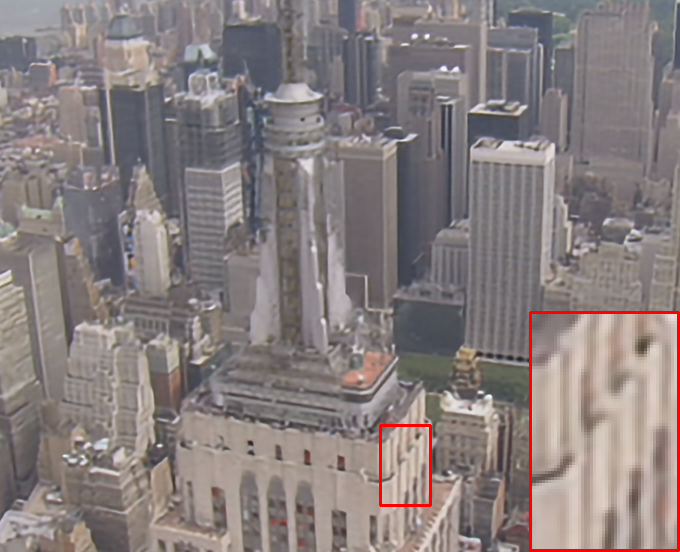}}
    }\vspace*{-2mm}
    \caption{        
    Examples of a reconstructed image from Videoset4 using different warping methods. FN2-SD stands for FlowNet2-SD. Clearly using JUBW yields sharper and more accurate reconstruction of the estimated frames compared to SPMC-FW~\cite{tao17} and the best VSR~\cite{kapp16} result.
    } 
    \label{fig:qualitative_warping}
    \vspace*{-5mm}
\end{figure}

\pagebreak 
    
Results are given in Table~\ref{tab:warping_results}. First, we note that our feed-forward implementation of FlowNet2-SD with SPMC-ED, which simply stacks frames and does not include an LSTM unit, outperforms the original recurrent implementation from Tao~\etal\cite{tao17}. Second, we see that our proposed JUBW operation generally outperforms SPMC-FW. We again performed experiments where we excluded temporal information, by inputting zero flows and duplicates of the center image. We now observe that including temporal information yields large improvements and increases the PSNR by $0.5$ to $0.9$. 
In contrast to the previous sections, we see such increase also for the Myanmar dataset. 
This shows that the proposed motion compensation can also exploit small motion vectors. The qualitative results in Fig.~\ref{fig:qualitative_warping} confirm these findings. 

Including the sub-pixel distance outputs from JUBW layer to enable better interpolation to the network leads to a smaller improvement than expected. Notably, without these distances the JUBW operation degrades to a simple nearest neighbor upsampling and nearest neighbor warping, but it still outperforms SPMC-FW. We conclude from this that one should generally avoid any kind of interpolation and leave it to the network. 
Finally, fine-tuning FlowNet2 on the video super-resolution task decreases the PSNR in some cases and does not provide the best results. We conjecture that this is due to the nature of optimization of the gradient through the warping operation, which is based on the reconstruction error and is prone to local minima.

    \section{Conclusions}
    In this paper, we performed an evaluation of different video super-resolution approaches using CNNs including motion compensation.  
    We found that the common practice of patch-based training and upsampling and warping separately yields almost no improvement when comparing the video super-resolution setting against the single-image setting. We obtained a significant improvement over prior work by replacing the patch-based approach by a network that analyzes the whole image. 
    As a remedy for the lacking standard motion compensation, we proposed a joint upsampling and backward warping operation and combined it with FlowNet2-SD~\cite{ilg17} and the SPMC-ED~\cite{tao17} architecture. This combination outperforms all previous work on video super-resolution. In conclusion, our results show that:
    1.) we can achieve the same or better performance with a formulation as a feed-forward instead of a recurrent network; 
    2.) performing joint upsampling and backward warping with no interpolation outperforms joint upsampling and forward warping and the common backward warping with interpolation; 
    3.) including sub-pixel distances yields a small additional improvement; and
    4.) joint training with FlowNet2-SD so far does not lead to consistent improvements and we leave a more detailed analysis     to future work.     
    
    \section*{Acknowledgements}
    We  acknowledge the DFG Grant BR-3815/7-1.

    \bibliographystyle{splncs03}
	\bibliography{egbib}

\end{document}


\pagestyle{headings}
\mainmatter

\def\GCPR17SubNumber{21}

    \title{Supplemental Material for \\ End-to-End Learning of Video Super-Resolution with Motion Compensation}


	\titlerunning{Supplemental Material}
	\authorrunning{Osama Makansi, Eddy Ilg and Thomas Brox}
	\author{Osama Makansi, Eddy Ilg, and Thomas Brox}
	\institute{Department of Computer Science, University of Freiburg}

	\maketitle

    \section{Computation of PSNR values}
    
    For all our evaluation results, the reported PSNR values are computed using only the Y channel of the estimated YCbCr image. In case of RGB images, we first convert to YCbCr color space and then compute on the Y channel.  For all experiments using the SRCNN~\cite{dong16} or VSR~\cite{kapp16} architecture, we follow ~\cite{kapp16} and for technical reasons crop away 12 pixels of the boundary from the estimated high-resolution images before computing PSNR values.     

    \section{Displacement magnitudes}
    
    We have noted that improvements using motion compensation are generally smaller on Myanmar than on Videoset4. In Table~\ref{tab:disp_mags}, we compute the average motion magnitudes of the datasets and note that the displacements are also generally smaller in the Myanmar validation set. 
    \begin{table}
        \centering
        \begin{tabular}{|c|c|}
            \hline
             Dataset &  Avg. Mag. \\
            \hline
            \hline 
            Myanmar training & $1.50$px\\
            Myanmar validation & $0.43$px \\ 
            Videoset4 & $1.29$px  \\
            \hline
        \end{tabular}
        \vspace*{2mm}
        \caption{Average motion magnitudes computed using FlowNet2~\cite{ilg17}. The numbers show that the Myanmar validation set has the smallest displacements.}
        \label{tab:disp_mags}
    \end{table}

    \section{Video super-resolution with patch-based training} 

    \newcolumntype{C}{>{\centering\arraybackslash}p{1.5cm}}

    Using patch-based training, we retrain and evaluate SRCNN~\cite{dong16} and VSR~\cite{kapp16} using different kind of motion compensations. However, the resulting PSNR scores in Table~\ref{tab:retrain_results} are all similar and we conclude that motion compensation on Myanmar has no effect. We also evaluate on Videoset4 (Table~\ref{tab:patch_videoset4}) and there see a small increment of $0.18$ for FlowNet2~\cite{ilg17} and FlowNet2-SD~\cite{ilg17}.
    
    \begin{table}     
        \begin{center}        
            \resizebox{\linewidth}{!}{
                \begin{tabular}{|c|@{}p{2pt}@{}|c|@{}p{2pt}@{}|C|C|C|C|C|} 
                    
                    \cline{1-1}\cline{3-3}\cline{5-9}                    
                    \multirow{2}{*}{Arch.} && 
                    \multirow{2}{*}{\backslashbox{Trained on}{Tested on}}  &&  
                    only & 
                    no & 
                    Drulea & 
                    FlowNet2- & 
                    FlowNet2 \\
                    && 
                    &&  
                    center &
                    warp & 
                    \cite{drulea11} & 
                    SD \cite{ilg17} & 
                    \cite{ilg17} \\
                    
                    \cline{1-1}\cline{3-3}\cline{5-9}
                    \multicolumn{8}{c}{} \\[-0.8\normalbaselineskip]
                    \cline{1-1}\cline{3-3}\cline{5-9}                    
                    SRCNN && only center &&  $\textbf{31.62}$ & 
                    - & - & - & - \\
                    \cline{1-1}\cline{3-3}\cline{5-9}
                    \multicolumn{8}{c}{} \\[-0.8\normalbaselineskip]                    
                    \cline{1-1}\cline{3-3}\cline{5-9}
                    
                    \multirow{5}{*}{VSR}
                    && only center &&
                    $\textbf{31.76}$ &  
                    - &
                    - &
                    - &
                    - \\
                    \cline{3-3}\cline{5-9}

                    && no warp && 
                    $31.80$ &  
                    $\textbf{31.83}$ &  
                    - &
                    - &
                    - \\
                    \cline{3-3}\cline{5-9}

                    && Drulea \cite{drulea11} && 
                    $31.77$ &  
                    $31.74$ &  
                    $\textbf{31.81}$ &  
                    - &
                    - \\
                    \cline{3-3}\cline{5-9}
                    
                    && FlowNet2-SD \cite{ilg17} && 
                    $31.75$ &  
                    $31.75$ &  
                    $31.79$ &  
                    $\textbf{31.77}$ &  
                    - \\
                    \cline{3-3}\cline{5-9}

                    && FlowNet2 \cite{ilg17} && 
                    $31.76$ &  
                    $31.76$ &  
                    $31.80$ &  
                    $31.78$ &  
                    $\textbf{31.79}$ \\ 
                    
                    \cline{1-1}\cline{3-3}\cline{5-9}
                    
                \end{tabular}
            }
            \vspace*{2mm}
            \captionof{table}{PSNR scores for patch-based video superresolution on the Myanmar validation set. We retrained the architecture of \cite{kapp16} using only the center frames (replicated three times), original images, and motion compensated frames. One can observe that all scores are nearly the same and motion compensation on the Myanmar validation set has no effect over providing original images or even only the center image.
            }
            \label{tab:retrain_results}
        \end{center}
    \end{table} 
        
    \begin{table} 
        \begin{center}
         \begin{tabular}{|c|c|} 
         \hline
         Motion compensation  & Videoset4 \\
         during training and testing     & PSNR \\
         \hline
         \hline
         only center & $24.60$ \\
         \hline
         no warp  & $24.59$ \\
         \hline
         Drulea \cite{drulea11} & $24.69$ \\
         \hline
         FlowNet2-SD \cite{ilg17} & $24.77$ \\
         \hline
         FlowNet2 \cite{ilg17} & $24.77$ \\
         \hline
        \end{tabular}
        \vspace*{2mm}
        \caption{Evaluation of the different retrained VSR models from Table~\ref{tab:retrain_results} on Videoset4. Motion compensation shows a small performance improvement.}
        \label{tab:patch_videoset4}
        \end{center}
    \end{table}

    \begin{table}
        \begin{center}
         \begin{tabular}{|l|c|c|} 
         \hline
         \multicolumn{1}{|c|}{Setting} & Patch-based & Image-based\\
         \hline
         \hline
         Learning rate & $1e-05$ & $1e-05$ \\
         Learning rate policy & fixed & multistep$^\dagger$ \\
         Momentum & $0.9$ & $0.9$ \\
         Weight decay & $0.0005$ & $0.0004$ \\
         Batch size & $240$ & $2$ \\
         Input resolution & $36\times36$ & $960\times540$ \\
         Image pixels in batch & $311$k & $1$M \\
         Training iterations & $200$k & $300$k \\
         Training time & $7$ hours & $32$ hours \\
         \hline
        \end{tabular}        
        \vspace*{2mm}
        \caption{Different settings of patch- and image-based traing. Settings are very similar, except that the number of pixels and trainig time in image-based training are larger. Note that the number of pixels is also further boosted much more by sliding the convolutions from the VSR architecture over the entire images with a stride of one.
        $^\dagger$multiplied by $0.5$ every $50$k iterations.}
        \label{tab:setting} 
        \end{center}
    \end{table}

    \section{Video super-resolution with image-based training} 

    We perform the same set of experiments as in the last section for image-based training. Comparing Table~\ref{tab:retrain_results} to Table~\ref{tab:image_retrain_results}, we find that PSNRs are generally around 1 point higher. We also provide all the training settings in Table~\ref{tab:setting}. Image-based training in general processes more training data and sees a lot of similar data during training by sliding the convolutions over an entire image. Motion compensation on Myanmar (Table~\ref{tab:image_retrain_results}) still seams to have little effect, while motion compensation on Videoset4 does show better PSNR values (Table~\ref{tab:image_videoset4}).
 
    \begin{table}         
        \begin{center}        
            \resizebox{\linewidth}{!}{
                \begin{tabular}{|c|@{}p{2pt}@{}|c|@{}p{2pt}@{}|C|C|C|C|C|}  
                    \cline{1-1}\cline{3-3}\cline{5-9}                    
                    \multirow{2}{*}{Arch.} && 
                    \multirow{2}{*}{\backslashbox{Trained on}{Tested on}}  &&  
                    only & 
                    no & 
                    Drulea & 
                    FlowNet2- & 
                    FlowNet2 \\
                    && 
                    &&  
                    center &
                    warp & 
                    \cite{drulea11} & 
                    SD \cite{ilg17} & 
                    \cite{ilg17} \\
                    
                    \cline{1-1}\cline{3-3}\cline{5-9}
                    \multicolumn{8}{c}{} \\[-0.8\normalbaselineskip]
                    \cline{1-1}\cline{3-3}\cline{5-9}                    
                    
                    \multirow{5}{*}{VSR}
                    && only center  &&
                    $\textbf{32.41}$ &  
                    - &
                    - &
                    - &
                    - \\
                    \cline{3-3}\cline{5-9}

                    && no warp && 
                    $32.38$ &  
                    $\textbf{32.55}$ &  
                    - &
                    - &
                    - \\
                    \cline{3-3}\cline{5-9}

                    && Drulea \cite{drulea11} && 
                    $32.37$ &  
                    $32.26$ &  
                    $\textbf{32.60}$ &  
                    - &
                    - \\
                    \cline{3-3}\cline{5-9}
                    
                    && FlowNet2-SD \cite{ilg17} && 
                    $32.35$ &  
                    $32.37$ &  
                    $32.58$ &  
                    $\textbf{32.62}$ &  
                    - \\
                    \cline{3-3}\cline{5-9}

                    && FlowNet2 \cite{ilg17} && 
                    $32.37$ &  
                    $32.36$ &  
                    $32.61$ &  
                    $32.61$ &  
                    $\textbf{32.63}$ \\ 
                    
                    \cline{1-1}\cline{3-3}\cline{5-9}                    
                \end{tabular}
            }
            \vspace*{2mm}
            \captionof{table}{PSNR scores from  Myanmar validation (ours). We now train the architecture of \cite{kapp16} by applying it as a convolution over the complete images. We again evaluate using only the center frame, original images and differently motion compensated frames. One can observe that scores are significantly better compared to the patch-based training, but motion compensation on the Myanmar validation set still has negligible effect compared to training on original frames.  
            }
            \label{tab:image_retrain_results}
        \end{center}
    \end{table} 
    
    \begin{table} 
        \begin{center}
         \begin{tabular}{|c|c|} 
         \hline
         Training input  & PSNR \\
         \hline
         \hline
         only center & $24.66$ \\
         \hline
         no warp  & $24.79$ \\
         \hline
         Drulea \cite{drulea11} & $24.91$ \\
         \hline
         FlowNet2-SD \cite{ilg17} & $25.12$ \\
         \hline
         FlowNet2 \cite{ilg17} & $25.13$ \\
         \hline
        \end{tabular}
        \vspace*{2mm}
        \caption{Evaluation of the different image-based models on Videoset4. Motion compensation in this case also shows a performance improvement.}
        \label{tab:image_videoset4}
        \end{center}    
    \end{table}

    \section{Joint Training} 
    
\begin{figure} 
        \resizebox{\linewidth}{!}{
            \subfigure[Initialization\label{fig:joint_img_a}]{\includegraphics[width=0.33\linewidth]{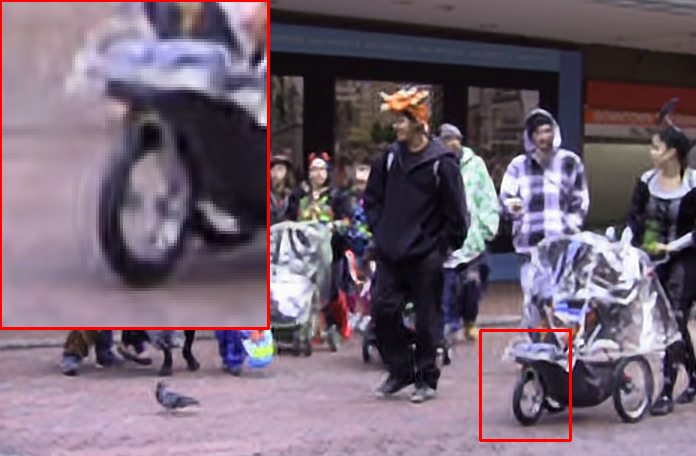}}
            \subfigure[After joint training\label{fig:joint_img_b}]{\includegraphics[width=0.33\linewidth]{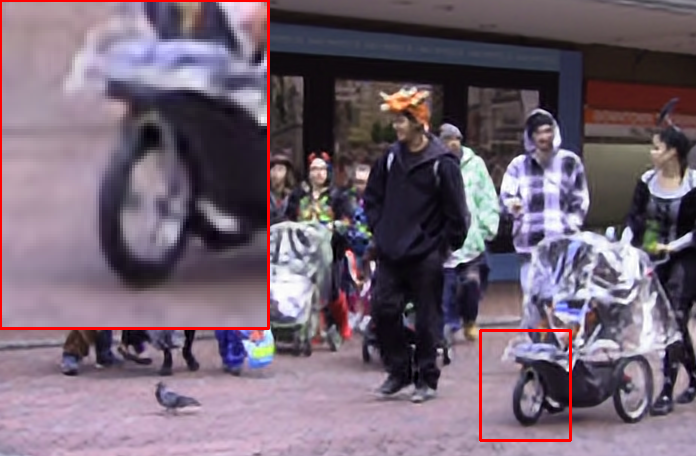}}
            \subfigure[After joint training with smoothness\label{fig:joint_img_c}]{\includegraphics[width=0.33\linewidth]{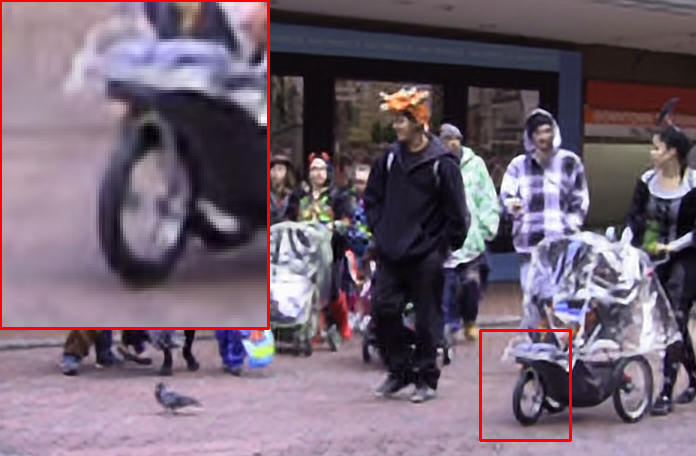}}
        }
        \resizebox{\linewidth}{!}{
            \subfigure[Initialization\label{fig:joint_flow_a}]{\includegraphics[width=0.33\linewidth]{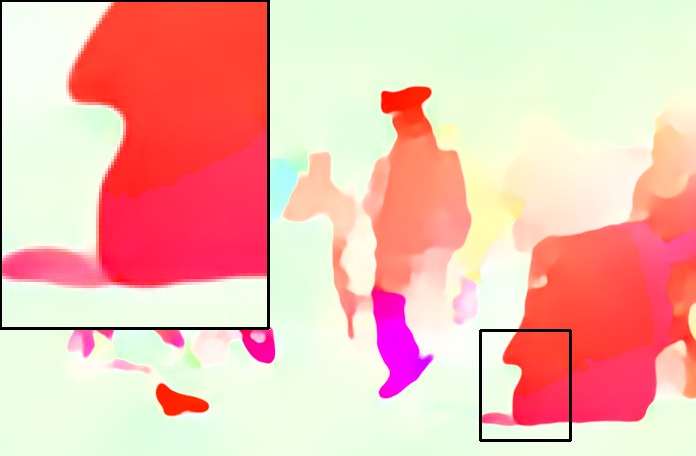}}
            \subfigure[After joint training\label{fig:joint_flow_b}]{\includegraphics[width=0.33\linewidth]{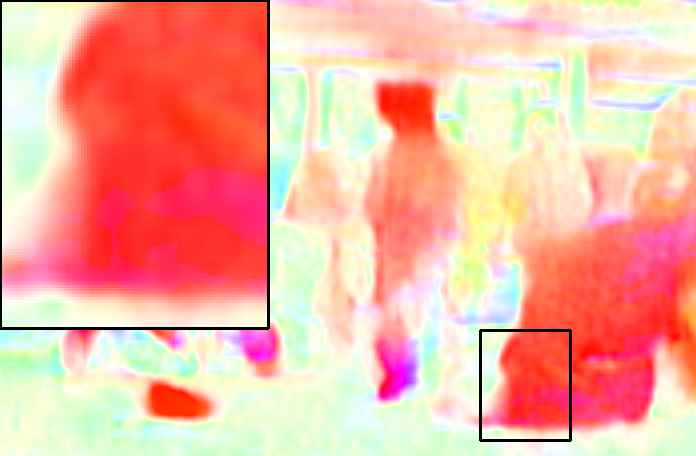}}
            \subfigure[After joint training with smoothness\label{fig:joint_flow_c}]{\includegraphics[width=0.33\linewidth]{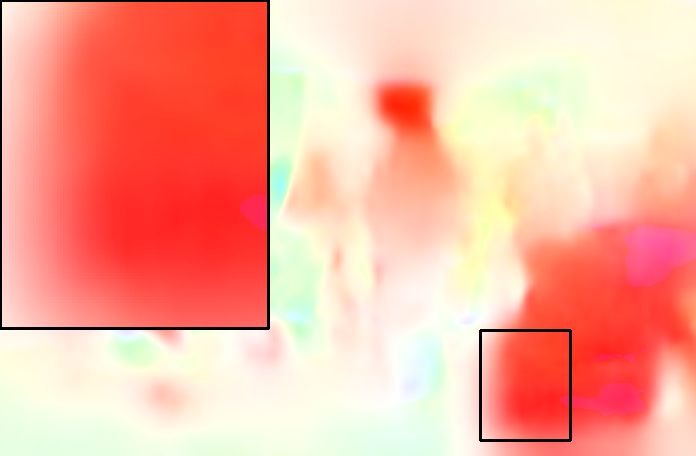}}
        }
        \caption{        
            Example super-resolved image after training FlowNet2-SD~\cite{ilg17} with VSR (a and d)  jointly (b and e) and including a smoothness constraint (c and f).
        } 
        \label{fig:joint}
    \end{figure}

    Since the FlowNet2-SD~\cite{ilg17} is completely trainable, we can refine the optical flow for the task of video super-resolution by training the whole network end-to-end with the super-resolution loss. This potentially allows the optical flow estimation to focus on aspects that are most relevant for the super-resolution task. 
    As an initialization we took the VSR network trained on FlowNet2-SD~\cite{ilg17} from the last section and used the same settings from Table~\ref{tab:setting}, but now cropped the images to a resolution of $256\times256$ to enable a batch size of $8$. We then trained for $100$k more iterations. 
    The result is given in Table~\ref{tab:joint_training} and Figures~\ref{fig:joint_img_b} and~\ref{fig:joint_flow_b}. We cannot see the flow itself improve, but we see a small improvement in the PSNR value on Videoset4 and from the images one can observe that the ringing artifacts disappear. 
    
    \begin{table}
        \begin{center}
         \begin{tabular}{|l|c|c|c|} 
         \hline
         \multicolumn{1}{|c|}{Test set} & After & After joint  & After joint  \\
                                        & initialization & training &   training with smoothness \\
         \hline
         \hline
         Myanmar validation & 32.62 & 32.63 & 32.61 \\
         \hline
         VideoSet4 & 25.12 & 25.21 & 25.19 \\
         \hline
        \end{tabular}        
        \vspace*{2mm}
        \caption{Evaluation of refining FlowNet2-SD~\cite{ilg17} on the super-resolution task.}
        \label{tab:joint_training} 
        \end{center}
    \end{table} 
    
    In Figure \ref{fig:joint_flow_b}, one can observe that many image details become flow artifacts. This is due to nature of the gradient through the warping operation; it corrects the flow vector to the best directly neighboring pixel, which is in most cases a local minimum. 
    Following \cite{godard16}, we add a regularization loss that penalizes deviations from smoothness  in the optical flow field, weighted with an image edge-aware term: 
    \begin{equation}
        \mathcal{L}_R = \sum_{i,j} \left(
          e^{-||\partial_x I_{i,j}||}\left(|\partial_x u| + |\partial_x v|\right)
        + e^{-||\partial_y I_{i,j}||}\left(|\partial_y u| + |\partial_y v|\right)\right) \mathrm{\,}
    \end{equation}
    where $I$ is the first image. $i,j$ is a pixel location and $u,v$ are the $x,y$ components of the flow vector. 
    The results of training with this additional smoothness term are given in Table~\ref{tab:joint_training} and Figures~\ref{fig:joint_img_c} and~\ref{fig:joint_flow_c}. The flow shows less artifacts than Figure~\ref{fig:joint_flow_b}, but compared to Figure~\ref{fig:joint_img_b} some very slight ringing artifacts still remain.  
    
 \section{Evaluating Architectures and Datasets\label{sec:arch_eval}} 


    In first part of the paper, the architecture from Dong~\etal~\cite{dong16} adapted to video super-resolution by Kappeler~\etal~\cite{kapp16} and the Myanmar training dataset were used. Here, we investigate the effect of architectures and training datasets. We extended the Myanmar training set by more high-resolution videos that we downloaded from Youtube. The resulting dataset has 162k frames of resolution $960\times540$ and we named it MYT. We evaluate and compare the SRCNN~\cite{dong16}, the FlowNet2-SD~\cite{ilg17} (here used for super-resolution, not flow) and the \mbox{encoder-/decoder} part of the architecture from Tao~\etal~\cite{tao17} (SPMC-ED) for single image super-resolution on the old and new datasets. The results are given in Table~\ref{tab:arch_results}. 
    
    \begin{table}
        \begin{center}
                \resizebox{\linewidth}{!}{
                    \begin{tabular}{|l|@{}p{2pt}@{}|c|c|c|c|}
                        \cline{1-1}\cline{3-6}
                                && SRCNN \cite{dong16} trained on & SRCNN \cite{dong16} & FlowNet2-SD~\cite{ilg17} & SPMC-ED~\cite{tao17} \\
                                && Myanmar training (ours) & trained on MYT & trained on MYT & trained on MYT \\
                        \cline{1-1}\cline{3-6}
                        \multicolumn{6}{c}{} \\[-0.8\normalbaselineskip]
                        \cline{1-1}\cline{3-6}
                        Myanmar validation (ours) && $32.42$ & $31.98$ & $31.47$ & $32.63$ \\
                        Videoset4                 && $24.63$ & $24.70$ & $24.93$ & $25.07$ \\
                        \cline{1-1}\cline{3-6}
                        Number of parameters & & 57K & 57K & 14M & 491K \\
                        \cline{1-1}\cline{3-6}
                    \end{tabular}
                }
                \vspace*{2mm}
                \caption{
                    PSNR values for different architectures and training datasets tested for single-image super-resolution. 
                }
                \label{tab:arch_results}
        \end{center}
    \end{table} 
    
    One can observe that SRCNN~\cite{dong16} tends to overfit on the Myanmar dataset. The much deeper FlowNet2-SD~\cite{ilg17} architecture performs worse on Myanmar, but can generalize better to Videoset4. The size of SPMC-ED~\cite{tao17} is between the former two and we observe that it performs best on Myanmar and also for generalization to Videoset4. It clearly gives better results than SRCNN~\cite{dong16} and for this reason we also use it for the final network in the paper.

	\bibliographystyle{splncs03}
	\bibliography{egbib}